\documentclass{article} % For LaTeX2e
\usepackage{iclr2024_conference,times}

\usepackage{amsmath,amsfonts,bm}

\def\eqref#1{equation~\ref{#1}}
\def\1{\bm{1}}

\DeclareMathAlphabet{\mathsfit}{\encodingdefault}{\sfdefault}{m}{sl}
\SetMathAlphabet{\mathsfit}{bold}{\encodingdefault}{\sfdefault}{bx}{n}

\usepackage{listings}
\usepackage{hyperref}
\usepackage{fontawesome}
\usepackage{url}
\usepackage{subcaption}
\usepackage{graphicx}
\usepackage{multirow}
\usepackage{booktabs}
\usepackage{amsmath}
\usepackage[noend]{algorithm2e}

\usepackage{algorithm,algorithmic}
\title{Inferflow: an Efficient and Highly Configurable Inference Engine for Large Language Models}

\iclrfinalcopy
\author{
\centerline{Shuming Shi${^1}$ \quad Enbo Zhao${^1}$ \quad Deng Cai${^1}$ \quad Leyang Cui${^1}$ \quad Xinting Huang${^1}$ \quad Huayang Li${^{1,2}} \thanks{Work was done during the internship at Tencent AI Lab.}$} \\
\\
\centerline{$^1$Tencent AI Lab} \\
\centerline{$^2$Nara Institute of Science and Technology}\\
 \centerline{{\faGithub}~\url{https://github.com/inferflow/inferflow}}
}

\begin{document}
\maketitle

\begin{abstract}
% Large language models (LLMs) have revolutionized the field of natural language processing, demonstrating unprecedented performance on a wide range of tasks. However, effectively utilizing these models in real-world applications remains a challenge due to their computational complexity and resource requirements. 
% In addition, the implementation also necessitates substantial expertise due to the diverse range of model architectures present in today's LLMs. To address these issues, 
We present Inferflow, an efficient and highly configurable inference engine for large language models (LLMs). With Inferflow, users can serve most of the common transformer models by simply modifying some lines in corresponding configuration files, without writing a single line of source code.
Compared with most existing inference engines, Inferflow has some key features. First, by implementing a modular framework of atomic build-blocks and technologies, Inferflow is compositionally generalizable to new models. Second, 3.5-bit quantization is introduced in Inferflow as a tradeoff between 3-bit and 4-bit quantization. Third, hybrid model partitioning for multi-GPU inference is introduced in Inferflow to better balance inference speed and throughput than the commonly-adopted partition-by-layer and partition-by-tensor strategies.
\end{abstract}

\section{Introduction}
Large language models (LLMs)~\citep{openai2023gpt4,llama2} have shown remarkable capabilities across a wide range of NLP tasks. 
Despite their impressive performance, deploying LLMs also poses several challenges, particularly in terms of computational complexity, resource requirements, and inference latency. 
The size of LLMs, often consisting of billions of parameters, makes them difficult to deploy in real-world applications, especially those that require low-latency and resource-constrained environments. 

In light of these challenges, we release Inferflow, an efficient and highly configurable inference engine for LLMs. We optimize the inference process by targeting on inference speed, throughput, result quality, VRAM consumption, and extensibility.

% , particularly in terms of computational complexity, resource requirements, and inference time. 

% Large language models (LLMs) have revolutionized the field of natural language processing, demonstrating unprecedented performance on a wide range of tasks. However, effectively utilizing these models in real-world applications remains a challenge due to their computational complexity and resource requirements. 
% In addition, decoding LLMs requires a high learning cost (e.g., model structure adaptation).

\begin{figure}
    \centering
    \includegraphics[width=\textwidth]{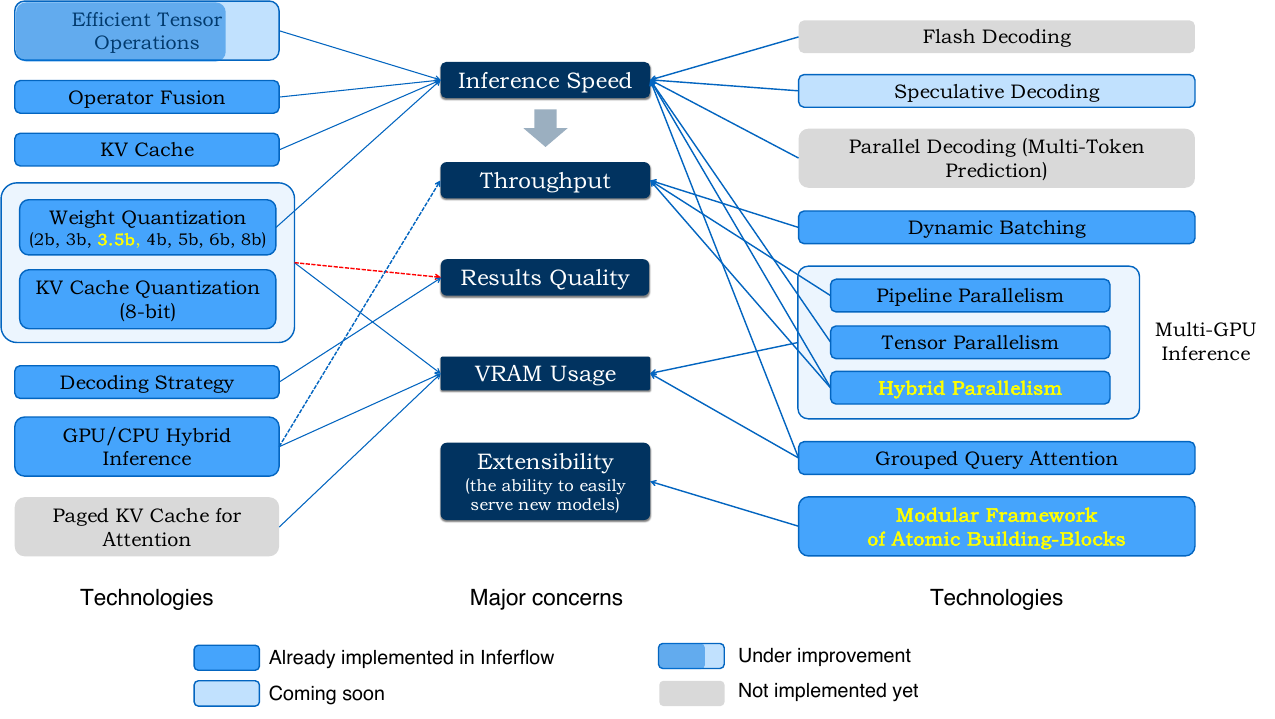}
    \caption{Implementation status of key technologies in Inferflow.}
    \label{fig:intro}
\end{figure}

Figure~\ref{fig:intro} lists major requirements for an LLM inference engine and possible technologies to address them.
The implementation status of the technologies in Inferflow is also depicted in the figure.

% computational efficiency and resource constraints.

Compared with existing popular inference engines~\citep{llama.cpp,DeepSpeed-MII,TensorRT-LLM,vLLM}, Inferflow has the following features:
\begin{enumerate}
\item \textbf{Extensible and highly configurable} (Section~\ref{sec:customized_module}): A typical way of using Inferflow to serve a new model is editing a model specification file, but not adding/editing source codes. We implement in Inferflow a modular framework of atomic building-blocks and technologies, making it compositionally generalizable to new models. A new model can be served by Inferflow if the atomic building-blocks and technologies in this model have been ``known'' to Inferflow.
\item \textbf{3.5-bit quantization} (Section~\ref{sec:3.5bit_quan}): Inferflow implements 2-bit, 3-bit, 3.5-bit, 4-bit, 5-bit, 6-bit and 8-bit quantization. Among the quantization schemes, 3.5-bit quantization is a new one introduced by Inferflow.
\item \textbf{Hybrid model partition for multi-GPU inference} (Section~\ref{sec:hybrid}): Inferflow supports multi-GPU inference with three model partitioning strategies to choose from: partition-by-layer, partition-by-tensor, and hybrid partitioning. Among them, hybrid partitioning is a new strategy introduced in this report.
\item \textbf{Wide file format support} (and safely loading pickle data): Inferflow supports loading models of multiple file formats directly, without reliance on an external converter. Supported formats include pickle, safetensors, llama.cpp gguf, etc. It is known that there are security issues to read pickle files using Python codes\footnote{https://huggingface.co/docs/hub/security-pickle}. By implementing a simplified pickle parser in C++, Inferflow supports safely loading models from pickle data.
\item \textbf{Wide network type support}: Supporting three types of transformer models: decoder-only models, encoder-only models, and encoder-decoder models.
\item \textbf{GPU/CPU hybrid inference}: Supporting GPU-only, CPU-only, and GPU/CPU hybrid inference.
% \item Many key modules of the model network can be specified by configration, including layer normalization functions, activation functions, position embedding algorithms, tensor names, etc.
\end{enumerate}

In this technical report, we give a brief description about the implementation of some key technologies in Inferflow for an efficient and extensible inference engine.
The source codes of Inferflow can be found at \url{https://github.com/inferflow/inferflow}.
%In this technicla report, we also introduce the various decoding strategies (Section~\ref{sec:decoding}) and grouped-query attention (Section~\ref{sec:group}) supported by Inferflow.

\section{Modular Framework of Atomic Build-Blocks}
\label{sec:customized_module}
Existing LLM inference engines typically adopt two approaches to support multiple models. The first way is adding ``if-else" clauses in multiple places of the source codes and implementing model-specific logic in corresponding code sections. The second way is maintaining a dedicated file for each model. Taking the Huggingface Transformers~\citep{transformers} as an example, the structure of Llama is defined in \textsc{models/llama/modeling\_llama.py}.
With the two approaches, supporting a new model often lead to the adding of new codes.
The second approach should be beneficial for exploring new training and inference techniques. However, if a new model is a combination of existing techniques, probably there are better ways to organize an inference engine to support the new model with less effort.
%Although these frameworks have pre-defined most of the mainstream models, when encountering new models without pre-defined model structure files, it is still necessary to manually modify these complex model codes. Some structures even need to access more underlying codes, for example, the activation function of huggingface transformer is defined in \text{transformers/activations.py}.

The design goal of Inferflow is to support a new model with zero code change, as long as the model is a combination of known techniques. The basic idea is to implement a modular framework of atomic building-blocks and technologies. The ``if-else" or ``switch" clauses are performed on the level of atomic building-blocks (or called modules) but not model types.
Examples of atomic modules include normalization functions, activation functions, position embedding algorithms, the grouped-query attention mechanism, etc.
In this way, a model is defined by determining the value or parameters of each atomic module. Naturally, this can be done with the aid of a configuration file.
By implementing such a modular framework, a typical way of serving a new model in Inferflow is editing a model specification file, but not adding/editing source codes.
Table~\ref{tab:custimzed_block} shows the contents of three model specification files (corresponding to three models) for comparison.
%To address the growing need for a flexible and efficient method to utilize various models, we propose a customized module that enables users to perform inference on different model structures by only modifying a configuration file.
%As shown in Table~\ref{tab:custimzed_block}, Inferflow can format a variety of model structures as a config file. It facilitates the easy adoption of different neural network structures, such as layer normalization functions, activation functions, and position embedding algorithms.

\begin{table}[h]
\centering
\resizebox{1.0\columnwidth}{!}{
\begin{tabular}{llll}
\toprule
\multirow{2}{*}{Model} & Mistral-7B-Instruct & Facebook-m2m100-418M  & BERT-base-multilingual \\
& \cite{jiang2023mistral} & \cite{fan2020englishcentric} & \cite{bert} \\
\midrule
model\_file\_format & pickle & pickle & safetensors \\
tokenizer\_file & tokenizer.json & vocab.json & tokenizer.json \\
tokenization\_algorithm & bpe & bpe & bpe \\
generation\_config & generation\_config.json  & generation\_config.json & generation\_config.json\\
\midrule
network\_structure: & & \\
\hspace{0.3cm}type & decoder\_only & encoder\_decoder & encoder\_only \\
\hspace{0.3cm}normalization\_function & rms & std & std \\
\hspace{0.3cm}activation\_function & silu & relu & gelu \\
\hspace{0.3cm}position\_embedding & rope & sinusoidal & empty \\
\hspace{0.3cm}qk\_column\_order & 2 & 2 & 2 \\
\hspace{0.3cm}qkv\_format & - & 1 & - \\
\hspace{0.3cm}tensor\_name\_prefix & model. & model. & bert.\\
\hspace{0.3cm}tensor\_name\_mapping & (1 line) & (16 lines) & (26 lines) \\
\bottomrule
\end{tabular}
}
\caption{The specification of three models.}
\label{tab:custimzed_block}
\end{table}
% \begin{lstlisting}[language=json, caption=JSON Configuration]
% {
%     "config_file": "config.json",
%     "model_files": ["pytorch_model.bin.index.json", "pytorch_model-00001-of-00002.bin", "pytorch_model-00002-of-00002.bin"],
%     "model_file_format": "pickle",

%     "tokenizer_file": "tokenizer.json",
%     "tokenization_algorithm": "fmm",
%     "generation_config": "generation_config.json",

%     "network_structure":
%     {
%         "type": "transformer.llama",
%         "normalization_function": "rms",
%         "activation_function": "silu",
%         "position_embedding": "rope",

%         "qk_column_order": 2,
%         "qkv_format": 1,

%         "tensor_name_prefix": "model.",
%         "tensor_name_mapping":
%         {
%         }
%     }
% }
% \end{lstlisting}

\section{Quantization}
\label{sec:3.5bit_quan}
Quantization is a process that reduces the numerical precision of model weights and/or activations to lower precision data types, such as 8-bit or 4-bit integers, instead of using floating-point numbers (e.g., float32 or float16). The main benefits of quantization are:

\begin{itemize}
  \item Reduced memory usage: Lower precision data types require less memory, enabling the deployment of large models on memory-constrained devices.
  \item Faster inference: Quantized operations can be significantly faster than their floating-point counterparts, leading to reduced inference time.
\end{itemize}

In Inferflow, we focus on \textbf{post-training quantization} \citep{dettmers20218bit,dettmers2022gptint,yao2022zeroquant}, which quantizes the model after training. It involves converting the model's weights and activations to lower precision data types and then fine-tuning the quantization parameters to minimize the loss of accuracy. Like in llama.cpp, we implement fast block-level linear quantization algorithms in Inferflow. The process can be formulated as follows:

Let $\boldsymbol{w}$ be the weight vector in a tensor block of the original model, and $\boldsymbol{q}$ be the corresponding quantized weight vector. The quantization process maps $\boldsymbol{w}$ to $\boldsymbol{q}$, where each element in $\boldsymbol{q}$ is a k-bit integer. The mapping is defined as:

\begin{equation}
\boldsymbol{q} = \mathrm{Round}(\frac{\boldsymbol{w} - \min(\boldsymbol{w})}{\max(\boldsymbol{w}) - \min(\boldsymbol{w})} \times (2^k - 1))
\label{eq:quan_encoding}
\end{equation}

where $\min(\boldsymbol{w})$ and $\max(\boldsymbol{w})$ are the minimum and maximum elements in the block, respectively, and $\mathrm{Round}(\cdot)$ is the element-wise rounding function.

After quantization, the weights can be dequantized back to their original precision for inference:

\begin{equation}
\boldsymbol{w}' = \frac{\boldsymbol{q}}{2^k - 1} \times (\max(\boldsymbol{w}) - \min(\boldsymbol{w})) + \min(\boldsymbol{w})
\label{eq:dequantization}
\end{equation}

where $\boldsymbol{w}'$ is the dequantized weight. 

\paragraph{3.5-bit quantization.} In $k$-bit post-training quantization, the model's weights are quantized to $k$ bits, which significantly reduces the model's memory footprint and can speed up inference, especially on hardware that is optimized for low-precision computations.
The value of $k$ can be any integer, with common practice being 2, 3, 4, 5, 6, and 8. However, we empirically found that LLMs perform reasonably well with 4-bit quantization, but the performance is relatively poor with 3-bit quantization. Therefore, we propose 3.5-bit quantization, where 7 bits are used to represent two adjacent weights.

Equation~\ref{eq:quan_encoding} is modified to
\begin{equation}
\boldsymbol{q} = \mathrm{Round}(\frac{\boldsymbol{w} - \min(\boldsymbol{w})}{\max(\boldsymbol{w}) - \min(\boldsymbol{w})} \times 10)
\end{equation}
Then we use a 7-bit number $q$ to represent two adjacent weights ${\boldsymbol{q}_{2i}}$ and ${\boldsymbol{q}_{2i+1}}$:
\begin{equation}
    \boldsymbol{q} = {\boldsymbol{q}_{2i}} \times 11 + {\boldsymbol{q}_{2i+1}}
\end{equation}

During dequantization, the weights are first decomposed by:
\begin{equation}
\begin{split}
    \boldsymbol{q}_1 &= \lfloor \boldsymbol{q} / 11 \rfloor \\
    \boldsymbol{q}_2 &= \boldsymbol{q}\mod 11
\end{split}
\end{equation}
Then, $\boldsymbol{q}_1$ and $\boldsymbol{q}_2$ are fed into Equation~\ref{eq:dequantization} for dequantization.

\begin{table}[]
    \centering
    \begin{tabular}{c|ccc|ccc|ccc}
    \hline
            & \multicolumn{3}{c|}{4-bit} & \multicolumn{3}{c|}{3-bit} & 
         \multicolumn{3}{c}{3.5-bit}\\
         $w$ & $q$ & $w'$ & $\Delta$ 
         & $q$ & $w'$ & $\Delta$
         & $q$ & $w'$ & $\Delta$ \\
         \hline
         -1 & 0 & -1.000 & 0.000 & 0 & -1.000 & 0.000 & 0 & -1.000 & 0.000 \\
         -0.9 & 1 & -0.833 & 0.067 & 0 & -1.000 & 0.100 & 0 & -1.000 & 0.100 \\ 
         -0.6 & 2 & -0.667 & 0.067 & 1 & -0.643 & 0.043 & 2 & -0.500 & 0.100 \\
         -0.4 & 4 & -0.333 & 0.067 & 2 & -0.286 & 0.086 & 2 & -0.500 & 0.100 \\
         -0.2 & 5 & -0.167 & 0.033 & 2 & -0.286 & 0.086 & 3 & -0.250 & 0.050 \\
         0 & 6 & 0.000 & 0.000 & 3 & 0.071 & 0.071 & 4 & 0.000 & 0.000 \\
         0.1 & 7 & 0.167 & 0.067 & 3 & 0.071 & 0.029 & 4 & 0.000 & 0.100 \\
         0.5 & 9 & 0.500 & 0.000 & 4 & 0.429 & 0.071 & 6 & 0.500 & 0.000 \\
         0.7 & 10 &0.667 & 0.033 & 5 & 0.786 & 0.086 & 7 & 0.750 & 0.050 \\
         1 & 12 & 1.000 & 0.000 & 6 & 1.143 & 0.143 &8 & 1.000 & 0.000 \\
         1.3 & 14 & 1.333 & 0.033 & 6 & 1.143 & 0.143 & 9 & 1.250 & 0.050 \\
         1.5 & 15 & 1.500 & 0.000 & 7 & 1.500 & 0.000 & 10 & 1.500 & 0.000 \\
         \hline
         Avg error & & & 0.031 & & & 0.075 & & & 0.046\\
         \hline

    \end{tabular}
    \caption{An example of quantization error statistics across various quantization bits.}
    \label{tab:exp_quan}
\end{table}

\begin{table}[]
    \centering
    \begin{tabular}{c|cccccccc}
    \toprule
        Quantization name & FP16 & Q8 & Q6 & Q5 & Q4\_B32 & Q4\_B64 & Q3H & Q3\_B32 \\
        Quantization bits & 16-bit & 8-bit & 6-bit & 5-bit & 4-bit & 4-bit & 3.5-bit & 3-bit \\
        Block size & - & 32/64 & 64 & 64 & 32 & 64 & 64 & 32 \\
        Actual bits/weight & 16 & 9/8.5 & 6.5 & 5.5 & 5 & 4.5 & 4 & 4 \\
    \midrule
        Bloom-3B & 18.056 & 18.056 & 18.095 & 18.169 & 18.304 & 18.557 & 19.208 & 19.882 \\
        LLAMA2-7B & 7.175 & 7.177 & 7.173 & 7.198 & 7.454 & 7.569 & 7.914 & 8.817 \\
        LLAMA2-13B & 6.383 & 6.382 & 6.391 & 6.434 & 6.514 & 6.647 & 6.897 & 7.278 \\
        Falcon-40B & 6.965 & 6.965 & 6.971 & 6.980 & 7.014 & 7.052 & 7.153 & 7.331 \\
    \bottomrule
    \end{tabular}
    \caption{Perplexity of some models across various quantization bits on Wikitext-2.}
    \label{tab:ppl}
\end{table}

Table~\ref{tab:exp_quan} presents the quantization errors across various quantization levels, using an example of 12 weights. It is evident that the proposed 3.5-bit quantization significantly reduces quantization errors compared to 3-bit quantization. We report, in Table~\ref{tab:ppl}, the perplexity of several models on Wikitext-2 for different quantization levels.
Several observations can be made: First, 8-bit quantization almost performs as well as the original FP16 version. Second, 6-bit quantization only leads to a tiny quality loss. Third, 3.5-bit quantization performs significantly better than 3-bit quantization. This justifies 3.5-bit quantization as an additional choice for users.
Please note that two FP16 numbers are stored for each block as auxiliary data for dequantization. As a result, the actual average bits per weight in memory is larger than quantization bits, as listed in the table.
It can be seen that Q3\_B32 (3-bit quantization with block size 32) has the same actual bits per weight as Q3H (3.5-bit quantization with block size 64). This observation means we should always adopt Q3H over Q3\_B32.
%The perplexity does not increase significantly as the bit quantization level decreases.

\section{Hybrid Partition}
\label{sec:hybrid}
For a large model, we may have to split it to multiple GPU devices because it exceeds the VRAM of a single GPU. Using multiple GPU devices to serve a model may also increase the inference speed.
%In order to accelerate the inference process and better utilize the computational resources, we can distribute the workload across multiple GPUs.
There are two common strategies for model partitioning in inference time: \textit{layer-wise partition} (pipeline parallelism) and \textit{tensor-wise partition} (tensor parallelism). In layer-wise partition, each GPU device is in charge of a subset of model layers. The GPU devices process the layers sequentially to complete the inference process.
%This approach improves throughput with the same inference speed.
On the other hand, tensor-wise partition involves dividing most tensors in the model into smaller segments, and then distributing these segments across multiple GPUs. Each GPU performs a partial computation, and the results are merged to obtain the final output. This method is particularly effective for the models with large weight matrices, as it reduces both the memory requirements and the computational load on each GPU. However, at each layer, the computation results on the GPUs have to be merged twice (for the self-attention sub-layer and the feed-forward sub-layer respectively). For machines with low inter-device communication bandwidth, the latency of merging the intermediate results may be high, especially when many GPU cards are involved in the merging process.

We propose and implement a new partitioning strategy called hybrid partitioning, with the goal of better balancing inference speed and throughput than the above two strategies.
An example of hybrid partitioning on 4 devices is shown in Table~\ref{tab:hybrid}.
Table~\ref{tab:hybrid_speed_compare} shows a preliminary experiment on 4 NVIDIA Tesla V100 cards for comparing the throughput and decoding speed of various partition strategies. The proposed hybrid partition surpasses the layer-wise partition in terms of decoding speed, and exceeds the tensor-wise partition regarding throughput.

\begin{table}[t]
    \centering
    \begin{tabular}{l|l}
    \toprule
        Device & Assignments  \\
    \midrule
       Device-0  & Layers: [1, 20], Heads: [1, 16] \\
       Device-1 & Layers: [1, 20], Heads: [17, 32] \\
       Device-2 & Layers: [21, 40], Heads: [1, 16] \\
       Device-3 & Layers: [21, 40], Heads: [17, 32]
 \\
    \bottomrule
    \end{tabular}
    \caption{An illustration of hybrid partition.}
    \label{tab:hybrid}
\end{table}

% todo

\begin{table}[t]
    \centering
    \begin{tabular}{c|cc}
    \toprule
        Partition  & Throughput (\#tokens/s) & Decoding speed (\#tokens/s) \\
    \midrule
        Layer-wise & 32 & 8 \\
        Tensor-wise & 12 & 12 \\
        Hybrid & 24 & 12 \\
    \bottomrule
    \end{tabular}
    \caption{Comparison of different partition strategies (model: Falcon-40B-Instruct, data type: FP16, device: Tesla V100, batch-size: 1)}
    \label{tab:hybrid_speed_compare}
\end{table}

\begin{figure}[t]
  \centering
  \begin{subfigure}[t]{0.4\textwidth}
    \includegraphics[width=\textwidth]{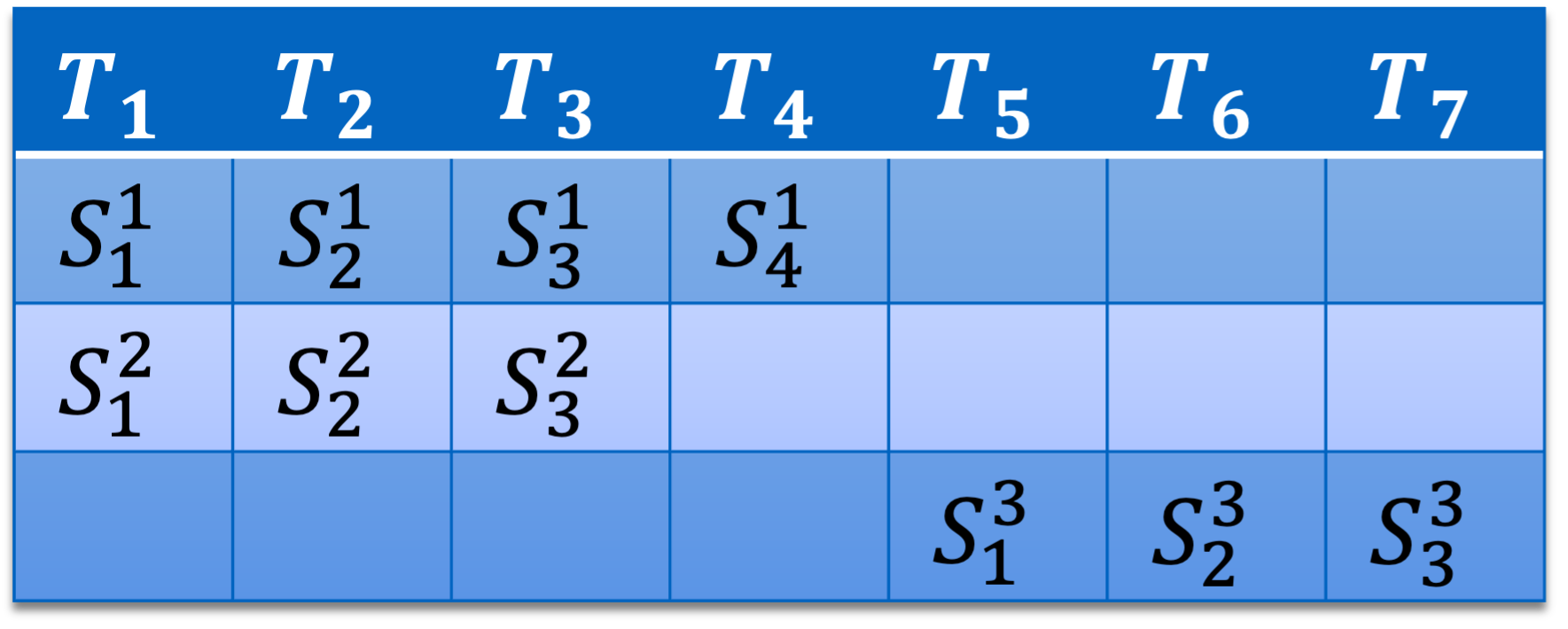}
    \caption{Static batching}
  \end{subfigure}
  \hfill
  \begin{subfigure}[t]{0.4\textwidth}
    \includegraphics[width=\textwidth]{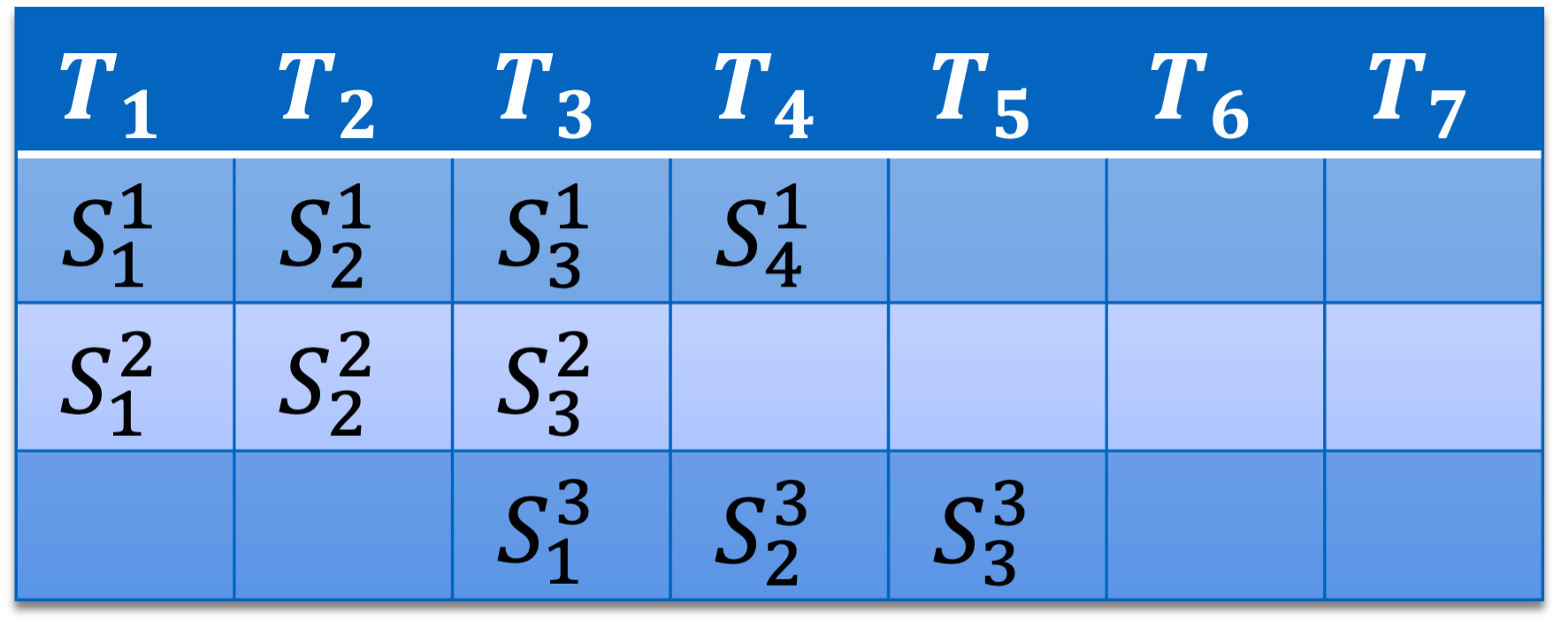}

    \caption{Dynamic batching\label{fig:dynamic_batching_illustration}}
  \end{subfigure}
  \caption{An illustration of two batching algorithms.\label{fig:batching_illustration}}
\end{figure}
\section{Dynamic Batching}
Batching is a technique that allows the model to process multiple input sequences simultaneously. In traditional static batching, input sequences of varying lengths are typically padded to the max length in each batch. This strategy is suitable for offline training and inference, since all input sequences can be first sorted based on their lengths. This step ensures that sequences with similar lengths are processed together, reducing the overall padding required.
While during online inference, it is common for different input sequences to be received and processed in real time. Inspired by \citet{orca}, we implement a dynamic batching technique that allows for generating responses without waiting for the decoding process of previous input sequences to be completed.
As illustrated in Figure~\ref{fig:batching_illustration}, consider two sentences $[[S^1_1, S^1_2, S^1_3, S^1_4], [S^2_1, S^2_2, S^2_3]]$ undergoing the decoding process. The system receives an additional request $S^3$ to generate at time step 3 ($T_3$). With static batching, the current batch must be completed before generating $S^3$ at $T_5$. In contrast, dynamic batching enables immediate inference of $S^3$ at $T_3$ by incorporating the sentence into the current inference batch.
Inferflow implements two member functions in the InferenceEngine class as the interface of dynamic batching:
\begin{itemize}
  \item AddQuery($S$): Add a new query ($S$) to the query pool
  \item Infer(): Conduct a single-step inference using the query pool, resulting in a set of next tokens.
\end{itemize}
Taking Figure~\ref{fig:dynamic_batching_illustration} as an example, a user call \textsc{AddQuery($S^3$)} at $T_3$, then call $\textsc{Infer()}$ to return $S^1_3, S^2_3, S^3_1$.

% 1. AddQuery: 向query pool添加一个query
% 2. Infer: 从query pool中取一个或者多个query，进行推理，对于每个query，生成next token

% which often leads to wasted computation on padding tokens.

% \begin{figure}
%   \centering
%   \begin{minipage}[b]{0.45\textwidth}
%     \includegraphics[width=\textwidth]{static_batching.png}
%     \caption{Static batching}
%   \end{minipage}
%   \hfill
%   \begin{minipage}[b]{0.45\textwidth}
%     \includegraphics[width=\textwidth]{dynamic_batching.png}
%     \caption{Dynamic batching}
%   \end{minipage}
% \end{figure}

\section{Decoding Strategies}
\label{sec:decoding}
Inferflow supports a diverse range of fast decoding strategies.

\paragraph{Top-$k$ Sampling}~\citep{fan-etal-2018-hierarchical} is used to ensure that the less probable words, which are in the unreliable tail of the distribution, should not have any chance to be selected. Only top-$k$ probable tokens are considered for generation.  

\paragraph{Top-$p$ Sampling}~\citep{Holtzman2020The} only considers the minimal set of most probable tokens $\mathcal{V}^p$ that cover a specified percentage $p$ of the distribution.

\paragraph{Frustratingly Simple Decoding (FSD)}~\citep{yang2023frustratingly} exploits the contrasts between the LLM and an auxiliary anti-LM constructed based on the current prefix. The anti-LM can be implemented as simple as an $n$-gram model. Specifically, the FSD score is defined as
 \begin{equation}
 \begin{split}
     \mathrm{FSD}(y|\mathbf{x}, \mathbf{y}_{< t}) = (1-\alpha)P_{\theta}(y|\mathbf{x}, \mathbf{y}_{< t})-  \alpha \times P_{\omega}(y|\mathbf{x}, \mathbf{y}_{< t})
\end{split}
\nonumber
 \end{equation}
where $P_\theta$ and $P_\omega$ represent the LM and the anti-LM respectively. The hyperparameter $\alpha \geq 0$ is used to balance the two scores. In practice, it first selects the top-$k$ most probable tokens according to $P_\theta (\cdot|\mathbf{x}, \mathbf{y}_{< t})$, denoted by $\mathcal{V}^{k}$. The token in $\mathcal{V}^{k}$ with the largest $\mathrm{FSD}$ score is chosen as the $t$-th token.

\paragraph{Randomized FSD} is a randomized variant of FSD. Note that FSD is a deterministic decoding methods and users may require a diverse set of outputs in many real-world applications. Randomized FSD select between from FSD and sampling in the first $T$ steps. In practice, we find that this strategy can produce diverse outputs without compromising the quality.

\paragraph{Temperature Sampling} is a decoding strategy to control the randomness in the sampling process. Instead of directly sampling tokens from the predicted distribution, temperature sampling introduces a hyperparameter ``temperature" $\tau$ that is used to adjust the skewness of the probability distribution:
\begin{equation}
    P(y |\mathbf{x}, \mathbf{y}_{<t}) = \frac{\exp(\mathbf{u}_y/\tau)}{\sum_j\exp(\mathbf{u}_j/\tau)}
    \nonumber
\end{equation}
where $\mathbf{u}_y$ is the logit of the token $y$ before softmax.

\paragraph{Typical Sampling} ~\citep{meister-etal-2023-locally} sorts the vocabulary according to the differences between distribution entropy and token probabilities. This process aims to generate output sequences with information content closely aligned to the model's expected information content, i.e., the conditional entropy. The candidate set $\mathcal{V}^c$ is a solution of the following problem:
\begin{equation}
\begin{split}
    \min_{\mathcal{V}^c}\sum_{y \in \mathcal{V}^c}|H(Y_t|&\mathbf{x},\mathbf{y}_{<t})+\log P(y|\mathbf{x},\mathbf{y}_{<t})| \\
    &\text{s.t.} \sum_{y \in \mathcal{V}^c}P(y|\mathbf{x},\mathbf{y}_{<t})\geq p 
\end{split}
\nonumber
\end{equation}

\paragraph{Mirostat Sampling}~\citep{basu2021mirostat} directly control the perplexity rate of the generated text. It firstly estimates the value
of $s$ assuming words follow Zipf’s law where $s$ is an exponent characterizing the distribution. Then it uses top-$k$ sampling to generate the new token where $k$ is a function of the estimated $s$ and of the target perplexity $\tau$ of the output text.

\paragraph{MinP Sampling} establishes a minimum probability threshold that a token must achieve to be included in the sampling process. This minimum value is calculated as a product of $\alpha$ and $p_{\text{max}}$, where $p_{\text{max}}$ is the top token's probability and $\alpha$  is a hyperparameter employed to manage the diversity of the sampling.

\paragraph{Tail Free Sampling (TFS)}~\citep{TFS} sorts token probabilities in descending order and truncates the tail of the distribution. Specifically, TFS calculates the first and second derivatives of the sorted distribution. A threshold $z$ is used to determine what part of the cumulative distribution of the second derivative weights to define the ``tail" of the distribution to be at. This is a hyperparameter that can be tuned to reduce the impact of most unlikely tokens.

\section{Grouped-query attention}
\label{sec:group}
Grouped-query attention \citep{ainslie-etal-2023-gqa} is proposed to reduce the memory consumption of the KV cache and to increase inference speed. This technique has been used in more and more LLMs, such as Llama2-70B, Mistral-7B \citep{jiang2023mistral}, and Falcon-40B \citep{falcon40b}.
Inferflow supports grouped-query attention as an atomic module. Grouped-query attention is automatically enabled when the value of hyper-parameter \textit{decoder\_kv\_heads} is smaller than another hyper-parameter \textit{decoder\_heads}.

Multi-head attention \citep{vaswani2023attention} splits the input into multiple heads, each responsible for attending to a different aspect of the input. The formulation of multi-head attention can be expressed as follows:

\begin{equation} \text{MultiHead}(Q, K, V) = \text{Concat}(\text{head}_1, \dots, \text{head}_h)W^O \end{equation}

where each head is computed as:

\begin{equation} 
\begin{split}
\text{head}_i &= \text{Attention}(Q_i,K_i,V_i) \\ 
Q_i &= QW_i^Q, K_i= KW_i^K, V_i=VW_i^V
\end{split}
\end{equation}

In this formulation, $Q_i$, $K_i$, and $V_i$ represent the query, key, and value matrices for the $i$-th head, and $W_i^Q, W_i^K, W_i^V$ and $W^o$ are the projection matrix. The Attention function can be any attention mechanism, such as scaled dot-product attention.

Despite their success, multi-head attention suffers from memory bandwidth in loading
keys and values. To reduce KV-cache, grouped-query attention constructs each group key and value head by mean-pooling all the original heads within that group. Assumed that we have $H$ attention heads with $G$ groups:
\begin{equation}
\begin{split}
    \text{group}_i &= \text{Attention}(Q_i,K^j,V^j) \\
    K^j &=\text{mean-pooling}(K^j_1,\dots,K^j_{H/G}) \\
        V^j &=\text{mean-pooling}(V^j_1,\dots,V^j_{H/G})
\end{split}
\end{equation}
If the number of groups is equal to the number of heads($H=G$), the grouped-query attention is reduced to multi-head attention.

\begin{algorithm}[h]
 \caption{Speculative Sampling}\label{alg:alg_sps}
 
\begin{algorithmic}

  \SetAlgoLined
  \SetKwProg{Def}{def}{:}{}

  \DontPrintSemicolon

    \STATE Input: number of lookahead step $K$, target sequence length $T$, target model $p(x)$, draft model $q(x)$, prefix input: $x_1,\dots,x_t$, whether accept based on sampling \textsc{is\_top\_sampling}

\STATE Initialise $n \leftarrow t$.
 
\WHILE{$n<T$} 
 \FOR{$t = 1: K$}
  \STATE Sample draft auto-regressively $\tilde x_{t} \sim p(x| , x_1, \dots, x_n, 
\tilde x_1, \dots, \tilde x_{t-1})$

\ENDFOR
\STATE In parallel, compute $K+1$ sets of logits from drafts $\tilde x_1, \dots, \tilde x_{K}$ :\[
q(x| , x_1, \dots, x_n),\
q(x| , x_1, \dots, x_n, \tilde x_1),\ \dots,\ q(x| , x_1, \dots, x_n, \tilde x_1, \dots, \tilde x_K)
\]
\FOR{$t = 1: K$} 
    \STATE  Sample $a\sim U[0,1]$ from a uniform distribution. 
        \STATE \textbf{if} $q(x|x_1,\dots,x_{n+t-1}) \ in \ \textsc{top\_sampling} (p(x|x_1,\dots,x_{n+t-1}))$ \AND \textsc{is\_top\_sampling}, \textbf{then}
        \STATE \quad Set $x_{n+t}\leftarrow \tilde x_t$ and $n\leftarrow n+1$.
        \STATE \textbf{elif} $a <\min\left(1, \frac{q(x|  x_1, \dots, x_{n+t-1})}{p(x|  x_1, \dots, x_{n+t-1})}\right)$, \textbf{then}
        \STATE \quad Set $x_{n+t}\leftarrow \tilde x_t$ and $n\leftarrow n+1$.
        \STATE \textbf{else}
        \STATE \quad sample $x_{n+t} \sim (q(x|  x_1, \dots, x_{n+t-1})-p(x|  x_1, \dots, x_{n+t-1}))_+$ and exit for loop.
        \STATE \textbf{end if}
\ENDFOR
\STATE If all tokens $x_{n+1}, \dots , x_{n+K}$ are accepted, sample an extra token $x_{n+K+1} \sim q(x| , x_1, \dots, x_n, x_{n+K})$ and set $n\leftarrow n+1$.
\ENDWHILE\label{alg:speculative_sampling}
\end{algorithmic}
\end{algorithm}

\section{Speculative Decoding}

To further enhance inference speed, Inferflow will soon incorporates speculative decoding \citep{chen2023accelerating,leviathan2022fast}, a technique that expedites the inference process for a large target LLM $p(·|x)$ by utilizing token proposals from a small draft model $q(·|x)$. The concept involves the draft model predicting the output $K$ steps ahead, while the target model dictates the number of tokens to accept from those speculations.

As shown in Algorithm~\ref{alg:speculative_sampling}, the draft model decodes $K$ tokens in the regular autoregressive fashion.
We get the probability outputs of the target and draft model on the new predicted sequence.
We compare the target and draft model probabilities to determine how many of the $K$ tokens we want to keep based on some rejection criteria. If a token is rejected, we resample it using a combination of the two distributions and do not accept any more tokens. If all $K$ tokens are accepted, we can sample an additional final token from the target model probability output.

The time complexity for the above algorithm is 
\begin{equation}
O(\frac{N}{r(K+1)} (t_{\text{draft}})K + t_{\text{target}})
\end{equation}
where $r$ is the acceptance rate indicating the percent of the draft tokens are accepted, which is the key factor of the speed-up ratio.
\citet{chen2023accelerating} accept the token by comparing a threshold $a\sim U[0,1]$ from a uniform distribution and the probability $\frac{q(x|  x_1, \dots, x_{n+t-1})}{p(x| x_1, \dots, x_{n+t-1})})$.
To further increase the acceptance rate $r$, we implement a flexible accept schema by consulting top-k / top-p decoding. If the draft token is present in the top-k / top-p pools, this token will be accepted by the target model.

\section{Conclusion}
We presented Inferflow, an efficient and highly configurable inference engine for LLMs. This technical report provided an overview of the key features and capabilities of Inferflow, briefly described the implementation of its key modules.

\section*{Acknowledgements}
Inferflow is inspired by the awesome projects of llama.cpp\footnote{\url{https://github.com/ggerganov/llama.cpp}} and llama2.c\footnote{\url{https://github.com/karpathy/llama2.c}}. The CPU inference part of Inferflow is based on the ggml library\footnote{\url{https://github.com/ggerganov/ggml}}. The FP16 data type in the CPU-only version of Inferflow is from the Half-precision floating-point library\footnote{\url{https://half.sourceforge.net/}}. We express our sincere gratitude to the maintainers and implementers of these source codes and tools.

% \begin{algorithm}[h]
%  \caption{推测解码 (Speculative Decoding)}\label{alg:alg_sps}
 
% \begin{algorithmic}

%   \SetAlgoLined
%   \SetKwProg{Def}{def}{:}{}

%   \DontPrintSemicolon

%     \STATE 输入: 每次推测解码的步数$K$, 目标序列长度$T$, 小模型$p(x)$, 大模型$q(x)$, 前缀文本 $x_1,\dots,x_t$，是否采用基于采样的接受策略\textsc{is\_top\_sampling}
% \STATE 初始化 $n \leftarrow t$。
% \WHILE{$n<T$} 
%  \FOR{$t = 1: K$}
%   \STATE  小模型自回归地草拟$K$个token $\tilde x_{t} \sim p(x| , x_1, \dots, x_n, 
% \tilde x_1, \dots, \tilde x_{t-1})$
% \ENDFOR
% \STATE 基于草拟的tokens$\tilde x_1, \dots, \tilde x_{K}$，并行计算大模型的分数（共$K+1$个）:
% $q(x| , x_1, \dots, x_n),\q(x| , x_1, \dots, x_n, \tilde x_1),\ \dots,\ q(x| , x_1, \dots, x_n, \tilde x_1, \dots, \tilde x_K)$

% \FOR{$t = 1: K$} 
%     \STATE  从均匀分布中采样一个随机数$a\sim U[0,1]$ 
%         \STATE \textbf{if} $q(x|x_1,\dots,x_{n+t-1}) \ in \ \textsc{top\_sampling} (p(x|x_1,\dots,x_{n+t-1}))$ \AND \textsc{is\_top\_sampling}, \textbf{then}
%         \STATE \quad Set $x_{n+t}\leftarrow \tilde x_t$ and $n\leftarrow n+1$.
%         \STATE \textbf{elif} $a <\min\left(1, \frac{q(x|  x_1, \dots, x_{n+t-1})}{p(x|  x_1, \dots, x_{n+t-1})}\right)$, \textbf{then}
%         \STATE \quad Set $x_{n+t}\leftarrow \tilde x_t$ and $n\leftarrow n+1$.
%         \STATE \textbf{else}
%         \STATE \quad 采样 $x_{n+t} \sim (q(x|  x_1, \dots, x_{n+t-1})-p(x|  x_1, \dots, x_{n+t-1}))$ 退出循环。
%         \STATE \textbf{end if}
% \ENDFOR
% \STATE 如果全部的token $x_{n+1}, \dots , x_{n+K}$ 被接受，额外采样一个token$x_{n+K+1} \sim q(x| , x_1, \dots, x_n, x_{n+K})$，并且设置$n\leftarrow n+1$。
% \ENDWHILE
% \end{algorithmic}
% \end{algorithm}

% \begin{CJK}{UTF8}{gbsn}
% 123

% \end{CJK}

\bibliography{iclr2024_conference}
\bibliographystyle{iclr2024_conference}

% \appendix
% \section{Appendix}
% You may include other additional sections here.

\end{document}